# The Role of AI in Human-AI Creative Writing for Hong Kong Secondary Students

Hengky Susanto [a, b], David James Woo [c], and Kai Guo [d]

[a] University of Massachusetts Lowell Department of Computer Science, USA

[b] Department of Science and Environmental Studies, The Education University of Hong Kong, Hong Kong, China

[c] Precious Blood Secondary School, Hong Kong, China

[d] Faculty of Education, The University of Hong Kong, Hong Kong, China

*Abstract* -- The recent advancement in Natural Language Processing (NLP) capability has led to the development of language models (e.g., ChatGPT) that is capable of generating human-like language. In this study, we explore how language models can be utilized to help the ideation aspect of creative writing. Our empirical findings show that language models play different roles in helping student writers to be more creative, such as the role of a collaborator, a provocateur, etc.

## 1. Introduction

Since the release of ChatGPT, this text generator application with artificial intelligence (AI) capability has taken the world by storm because of its ability to generate human-like text. For this reason, this application has raised many concerns among educators for its capability to facilitate academic dishonesty, such as students using ChatGPT to generate essays for assignments. However, at the same time, we also recognize its potential to be used for good, such as helping non-native English speakers to enhance creative writing in English.

## 2. Data Collection

To realize AI's capability to generate human-like text as a learning tool, we first built and prototyped AI-based sentence generator tools using free and open-source Natural Language Processing (NLP) based libraries available for the Python programing language, such as autoregressive language models (Radford et al., 2019) on which ChatGPT is based. In other words, our tools are capable of generating human-like, English language text at word, sentence and paragraph levels.

Importantly, we launched an English language creative writing contest for Hong Kong secondary school students. Prior to the competition, students were given a workshop session to learn about short story writing as well as training on how to use the tools. Then, participating students utilized the tools to facilitate idea generation for their creative writing. Each student wrote a short story using a mix of their own sentences and AI generated sentences together. At the end of the competition, we asked the participating students to complete a post-contest survey.

## 3. Analysis Methodology

We analyzed students' stories to address if and how the AI-based tools had helped the creative aspect of students in completing their task of writing a short story. To address these questions, we employed a framework comprising four categories of creativity proposed by Kaufman and Beghetto (2009). In this "four C" model: "big C" refers to a level creativity leading to extraordinary achievements (for example, winning a Nobel Prize); "pro-C" refers to creativity at a professional level (for example, designer, professional musician, journalist, etc.), "little-C" is everyday creativity (for example, cooking, fixing broken items, etc.) or hobbies, and "mini-C" is creativity that are useful for learning purposes (for example, identifying the rules and constraints of a given domain like math, music, writing, etc.). In our analysis, we focus on the "mini-C" in the context of AI text generator contributions to students' performance in creative writing.

At a fine-grain level, we analyzed the position where AI-generated sentences were placed within a basic paragraph structure, comprising a topic sentence, supporting details, and a concluding sentence. This analytical framework gave us specific information of how the AI-generated text has contributed to the student's writing process. For example, the first sentence in a paragraph is often used to state the main idea or topic of the paragraph. Thus, when a student decides to place the generated text in the beginning of a paragraph, this is an indication that the generated text is used to trigger new ideas to build the storyline. At a broader level, we looked at which paragraphs within a story use more AI-generated sentences.

### 4. Findings and Discussion

Through analysis of the student's writings from the competition, we observed that many of the students used AI-generated sentences for their opening sentence of the *first* paragraph. This is an important finding because this indicates the AI text generator is used to give them a quicker or guided head start. Secondly, AI-generated sentences are frequently found in the beginning of a paragraph. For example, in some short stories, a significant number of the paragraphs began with AI-generated sentences. From these observations, we may conclude that students find the tools most useful to trigger new ideas to develop the story further. Our analysis of post-contest surveys affirmed this conclusion. For example, the students found the tools helpful in providing them with ideas as they were writing their story. In sum, by using the tools to generate text for the beginning of the paragraphs, these students found the tools useful for invoking new ideas or to find inspiration to develop the story plot.

Furthermore, we investigated the role(s) of the tools in co-authoring the short story. To realize this, we analyzed the surveys to understand the nature of student interactions with the AI-based tools. We observed that some students only selected the sentence or words that fit her story narrative. These observations suggest that the AI text generator plays the role of a "partner" in helping them generate new ideas and develop the story, which is also an approach to be more creative (Osborn, 1963). One student mentioned in the survey that, "I chose the sentence generator to get ideas...to make my story more interesting." The student explored and "compared which tools would make the writing better." Similarly, some students found the tools to be useful in helping them to locate words that express their thoughts more accurately. Another student said, "First I would think of a single sentence about the writing which is incomplete, then I would put it into the generator and let it do its work." This student used the tools to "find the words which make the writing better."

On the other hand, the AI-based sentence generator tools may play the role of a provocateur, generating ideas which may not be directly or obviously aligned with the original storyline. A student found it challenging to integrate the sentences generated by the tools into her storyline. In the survey, this student wrote, "It's difficult to write with my own words and the AI's words because if you have the story planned, the AI might switch it up a bit, sometimes it might help give you new ideas but I have a strict plan with how I want the story to go…" In this case, the student collaboration with the AI text generator can be an experience of forced association (Osborn, 1963), which is a creative technique pairing random ideas together to provoke novel solutions. This technique is widely practiced to achieve innovative solutions to certain problems.

## 5. Conclusion

To conclude, our empirical findings indicate that students can utilize AI-based sentence generator tools like ChatGPT to trigger ideas to further develop a story. They also indicate that different students can utilize the text generator differently in helping them to be more creative in their writing, perhaps depending on the student's English language ability. For some students, the tools are more of a partner, while for other students the tools have played the role of a provocateur.

These insights are crucial for our future work of using the AI-based sentence generator tools for learning and teaching creative writing. We envision teachers can use these tools to specify the outputs from the language models to fit their teaching objectives and the desired learning outcomes. In other words, the teacher could customize the type and extent of help students get from the language models based on factors such as students' grade level, writing techniques, etc. Using tools in these ways may alleviate educators' concerns for AI-based sentence generator tools' capability to facilitate academic dishonesty.